\documentclass[10pt,twocolumn,letterpaper]{article}

\usepackage{iccv}
\usepackage{times}
\usepackage{epsfig}
\usepackage{graphicx}
\usepackage{amsmath}
\usepackage{amssymb}
\usepackage{booktabs}
\usepackage{enumitem}
\usepackage{subfigure}

\usepackage[breaklinks=true,bookmarks=false]{hyperref}

\iccvfinalcopy 

\def\iccvPaperID{****} 

\ificcvfinal\fi

\begin{document}

\title{Cross-View Policy Learning for Street Navigation}

\author{Ang Li\textsuperscript{1}\thanks{{Equal contribution}}\quad Huiyi Hu\textsuperscript{1}\footnotemark[1]\quad Piotr Mirowski\textsuperscript{2}\quad Mehrdad Farajtabar\textsuperscript{1}\\
\textsuperscript{1}DeepMind, Mountain View, CA\\
\textsuperscript{2}DeepMind, London, UK\\
{\tt\small \{anglili, clarahu, piotrmirowski, farajtabar\}@google.com}
}

\maketitle
\ificcvfinal\thispagestyle{empty}\fi

\begin{abstract}
The ability to navigate from visual observations in unfamiliar environments is a core component of intelligent agents and an ongoing challenge for Deep Reinforcement Learning (RL). Street View can be a sensible testbed for such RL agents, because it provides real-world photographic imagery at ground level, with diverse street appearances; it has been made into an interactive environment called StreetLearn and used for research on navigation. However, goal-driven street navigation agents have not so far been able to transfer to unseen areas without extensive retraining, and relying on simulation is not a scalable solution. Since aerial images are easily and globally accessible, we propose instead to train a multi-modal policy on ground and aerial views, then transfer the ground view policy to unseen (target) parts of the city by utilizing aerial view observations. Our core idea is to pair the ground view with an aerial view and to learn a joint policy that is transferable across views. We achieve this by learning a similar embedding space for both views, distilling the policy across views and dropping out visual modalities. We further reformulate the transfer learning paradigm into three stages: 1) cross-modal training, when the agent is initially trained on multiple city regions, 2) aerial view-only adaptation to a new area, when the agent is adapted to a held-out region using only the easily obtainable aerial view, and 3) ground view-only transfer, when the agent is tested on navigation tasks on unseen ground views, without aerial imagery. Experimental results suggest that the proposed cross-view policy learning enables better generalization of the agent and allows for more effective transfer to unseen environments.
\end{abstract}


\section{Introduction}

Stranded on Elephant Island after the shipwreck of the \emph{Endurance} expedition, Ernest Shackleton, Frank Worsley and their crew attempted, on 24 April 1916, a risky 720-mile open-boat journey to South Georgia. They had duly studied the trajectory using nautical maps, but the latter froze and became illegible. It is only through their extraordinary navigation skills, memory, and by transferring  knowledge derived from a top-view representation to visual and compass observations as they sailed, that they ultimately reached the shores of South Georgia two weeks later. Such a feat has been cited as a prime example of complex human spatial navigation in unknown environments~\cite{ekstrom2018human}: having gained expertise in navigating using both maps and sea-level observations, they could adapt to an unknown environment by studying maps and then transfer that knowledge on their new journey.

\begin{figure}
    \centering
    \includegraphics[width=\linewidth]{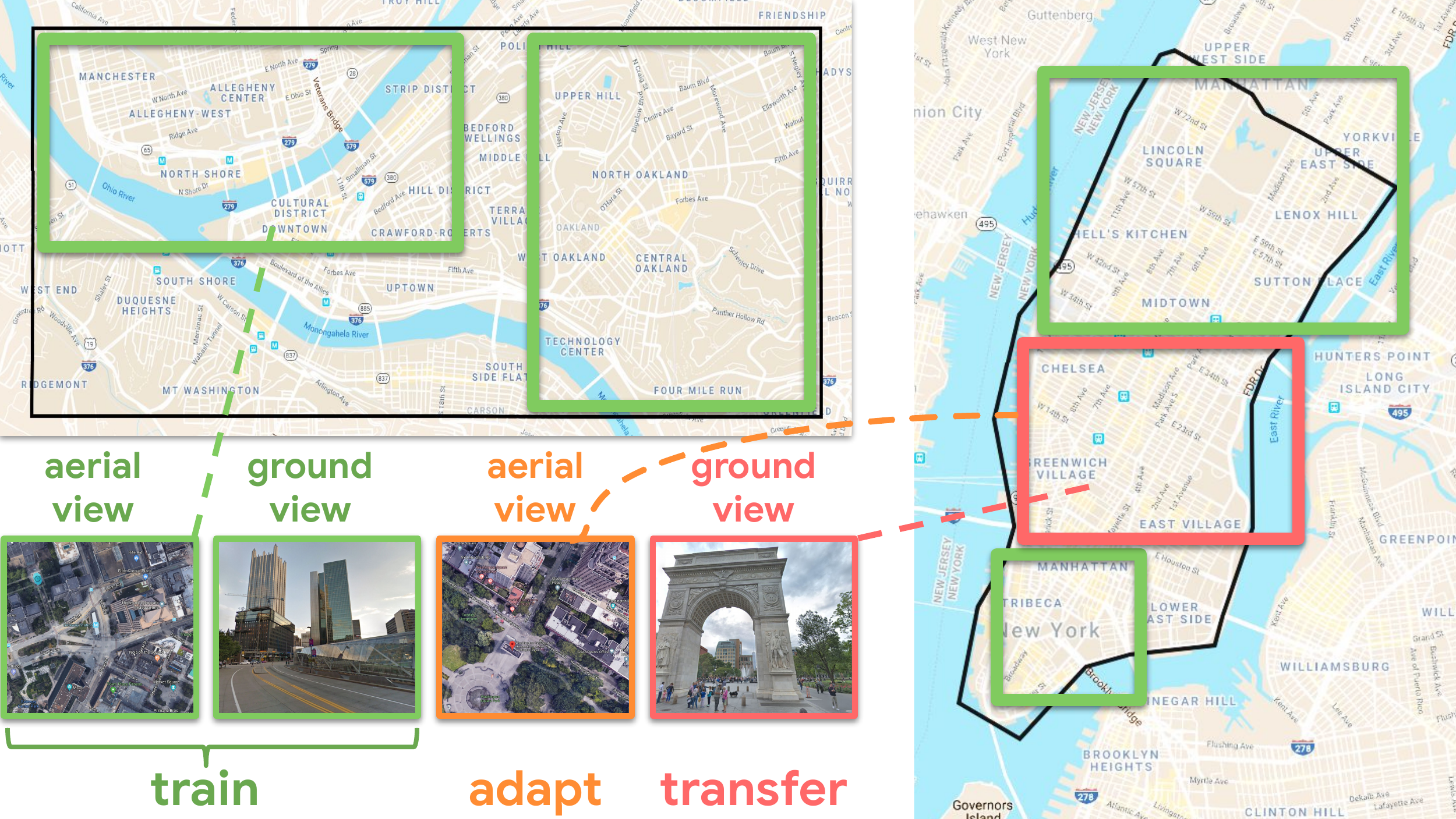}
    \caption{The street navigation agent observes both ground and aerial views in the training phase. The agent learns a view-invariant policy to associate the two views. Once the policy is learned, the agent becomes capable of continual training with interchangeable viewpoints. When being transferred to an unseen area, the agent is adapted  using only the aerial view observations, which are easily accessible. The agent is then transferred to the ground view environment (without access to aerial-view images) for testing. Images: Google Maps and Street View.}
    \label{fig:my_label}
\end{figure}



The ability to navigate in familiar and unfamiliar environments is a core component of animal and artificial intelligence. The research on artificial agent navigation can be applied to real world domains ranging from the neuroscience of grid and place cells in mammals~\cite{banino2018vector,cueva2018emergence} to the autonomy of indoor and outdoor mobile robots \cite{zhu2017target,tai2017virtual,mo2018adobeindoornav,shah2018airsim,wolcott2014visual,DBLP:journals/corr/abs-1805-06066}.

We focus on the visual navigation task that trains an agent to navigate in a specific area by using a single sensory modality, integrates visual perception and decision making processes, and typically does not rely on maps. A challenging question arises: \textit{how to efficiently transfer the agents to new or previously unseen areas?} In the absence of extra information, existing solutions typically require to retrain the agent on that unseen area, which is computationally expensive~\cite{chaplot2016transfer}.
Alternatively, one can simplify navigation tasks so as not to require local knowledge~\cite{zhu2017target} or to rely on additional navigation instructions~\cite{chen2018touchdown,Hermann2019LearningTF}. Generalization to unseen environments can be obtained by approaching navigation as a one-shot learning task with an auxiliary memory in simple and procedurally generated environments~\cite{wayne2018unsupervised,zhang2017neural} or by building complex simulators for more complex environments~\cite{mo2018adobeindoornav,tai2017virtual}. It is however expensive to build a simulator for offline retraining (especially in the case of unconstrained outdoor environments) and street-level images are expensive to collect as one has to drive everywhere to take panoramic photographs. As a consequence, enabling an agent to navigate in unseen locations, without fully retraining it from scratch, is still a challenging problem.

Inspired by the observation that humans can quickly adapt to a new city simply by reading a map, we explore the idea of incorporating comparable top-down visual information into the training procedure of navigation agents, in order to help them generalize to previously unseen streets. Instead of using a human-drawn map, we choose aerial imagery, as it is readily available around the world. 
Moreover, humans can easily do without maps once they become familiar with an environment. This human versatility motivates our work on training flexible RL agents that can perform using both first-person and top-down views.

We propose a novel solution to improve transfer learning for visual navigation in cities, leveraging easily accessible aerial images (Figure~\ref{fig:my_label}). These aerial images are collected for both source (training) and target (unseen or held-out) regions and they are paired with ground-level (street-level or first-person) views based on their geographical coordinates. We decompose the transfer task into three stages: \textit{training} on both ground-view and aerial-view observations in the source regions, \textit{adaptation} using only the aerial-view observations in the target region, and \textit{transfer} of the agent to the target area using only ground-view observations. Note that our goal remains to train agents to navigate from ground-view observations. The RL agent should therefore have access to the aerial views only during the first (training) and second (adaptation) stages, but not during the third (transfer) stage when it is deployed in the target area.

The gist of our solution is transfering the agent to an unseen area using an auxiliary environment built upon a different but easily accessible modality -- the aerial images. This requires the agent to be flexible at training time by relying on interchangeable observations. We propose a cross-view framework to learn a policy that is invariant to different viewpoints (ground view and aerial view). Learning view-invariant policy relies on three main ingredients: (a) an $L_2$ distance loss to minimize the embedding distance between the two views, (b) a dual pathway, each with its own policy, with a Kullback-Leibler (KL) loss on the policy logits to force these two policies to be similar, and (c) a dropout module called \textit{view dropout} that randomly chooses the policy logits from either view to select actions. The proposed architecture naturally works with interchangeable observations and is flexible for training with both views jointly or with only view at a time. This makes it a flexible model that can be shared across the three stages of transfer learning. 

We build our cross-view policy architecture by extending the RL agents proposed in \cite{streetlearn} into a two-stream model that corresponds to the two views. Our agents are composed of three modules: a convolutional network~\cite{lecun1998gradient} responsible for visual perception, a local recurrent neural network (RNN) or Long Short-Term Memory (LSTM) \cite{hochreiter1997long} for capturing location-specific features (\emph{locale} LSTM), and a policy RNN producing a distribution over the actions (\emph{policy} LSTM). 

We build our testbed, called \textit{StreetAir} (to the best of our knowledge, the first multi-view outdoor street environment), on top of \textit{StreetLearn}, an interactive first-person street environment built upon panoramic street-view photographs~\cite{mirowski2019streetlearn}. We evaluate it on the same task as in~\cite{streetlearn}, namely goal driven navigation or the \emph{courier} task, where the agent is only given the latitude and longitude coordinates of a goal destination, without ever being given its current position, and learns to both localize itself and plan a trajectory to the destination.  Our results suggest that the proposed method transfers agents to unseen regions with higher zero-shot rewards (transfer without training in the held-out ground-view environment) and better overall performance (continuously trained during transfer) compared to single-view (ground-view) agents.

\vskip 0.5em
\noindent\textbf{Contribution.}
 Our contributions are as follows.
\begin{enumerate}[leftmargin=*,topsep=0.5em,]
\itemsep 0pt
\item 
We propose to transfer the ground-view navigation task between areas by leveraging a paired environment based upon easily accessible aerial-view images. 
\item
We propose a cross-view policy learning framework to encourage transfer between observation modalities via both representation-level and policy-level associations, and a novel view dropout to force the agent to be flexible and to use ground and aerial views interchangeably.
\item
We propose a three-stage procedure as a general recipe for transfer learning: cross-modal training, adaptation using auxiliary modality, and transfer on  main modality.
\item We implement and evaluate our agents on \textit{StreetAir}, a realistic multi-view street navigation environment that extends \textit{StreetLearn}~\cite{mirowski2019streetlearn}.
\end{enumerate}

\section{Related Work}

\subsection{Visual Navigation}

Zhu \etal~\cite{zhu2017target} proposed an actor-critic model whose policy was a function of the goal as well as of the current state, both presented as images.
Subsequent work on Deep Reinforcement Learning focused on implicit goal-driven visual navigation~\cite{mirowski2016learning,chaplot2017gated,tai2017virtual,zhang2017deep} and addressed generalization in unseen environments through  implicit~\cite{oh2016control,wayne2018unsupervised} or explicit~\cite{zhang2017neural,parisotto2018global} map representations.
Gupta \etal~\cite{gupta2017unifying} introduced landmark- and map-based navigation using a spatial representation for path planning and a goal-driven closed-loop controller for
executing the plan. A successor-feature-based deep RL algorithm that can learn to transfer knowledge from previously mastered navigation tasks to new problem instances was proposed in~\cite{zhang2017deep}.
However, the above works either relied on simulators or attained navigation in simple, unrealistic, or limited indoor environments.

There has been a growing interest in building and benchmarking visual navigation using complex simulators~\cite{kolve2017ai2,shah2018airsim} or photorealistic indoor environments~\cite{mo2018adobeindoornav}. 
By contrast, we built our work on the top of a realistic environment \textit{StreetLearn}~\cite{streetlearn,mirowski2019streetlearn}, made from Google Street View imagery and Google Maps street connectivity.

\subsection{Cross-View Matching}
Matching street viewpoints with aerial imagery has been a challenging computer vision problem \cite{li14geolocation,Kim2017LearnedCF,Lin2015,zcanli2015GeolocalizationUV}. Recent approaches include geometry-based methods and deep learning. Li \etal would extract geometric structures on the ground between street and ortho view images, and measure the similarity between modalities by matching their linear structures \cite{li14geolocation}. Bansal \etal\cite{Bansal:2011:GSV:2072298.2071954} proposed to match lines on the building facades. Lin \etal proposed to learn a joint embedding space using a deep neural network between street views and aerial views \cite{Lin2015,Tian2017CrossViewIM}. All these works aim at utilizing cross-view matching to achieve image-based geolocation - specifically, finding the nearest neighbors, in some embedding space, between the query street image and all the geo-referenced aerial images in the database. Our work is closely related to cross-view matching, but instead of supervised learning, we study how cross-view learning could improve RL-based navigation tasks.

\subsection{Multimodal Learning}
Our work is also generally related to multimodal learning since street views and aerial views are not taken from the same type of cameras; they are basically from two different modalities. Many of the existing multimodal learning works focus on merging language and visual information.
In the visual navigation domain, Hermann \etal built upon the \textit{StreetLearn} environment~\cite{streetlearn} with additional inputs from language instructions, to train agents to navigate in a city by following textual directions \cite{Hermann2019LearningTF}.
Anderson \etal proposed the vision-and-language navigation (VLN) task based upon an indoor environment \cite{DBLP:conf/cvpr/AndersonWTB0S0G18}. 
Wang \etal \cite{wang2018rcm-sil} proposed to learn, from paired trajectories and instructions, a cross-modal critic that provides intrinsic rewards to the policy and utilizes self-supervised imitation learning.

\subsection{Knowledge Distillation}
Our work is related to Network Distillation~\cite{hinton2015distilling,ba2014deep} and its many extensions~\cite{mirzadeh2019improved,li2017learning,zhang2018deep,romero2014fitnets}, as one way to transfer knowledge. A \emph{student} network tries to indirectly learn from a \emph{teacher} network by imposing a Kullback-Leibler (KL) loss between its own and the teacher's softened logits, \textit{i.e.}, trying to mimic the teacher's behavior. 
In~\cite{gupta2016cross} Gupta \etal generalize knowledge distillation for two modalities (RGB and depth) at the final layer by minimizing the $L_2$ loss for object and action detection.
The hallucination network in~\cite{hoffman2016learning} was trained on an existing modality to regress the missing modality using $L_2$ loss, and leveraged multiple such losses for multiple tasks. This work has been extended by Garcia \etal by adding $L_2$ losses for reconstructing all layers of the depth network and a cross entropy distillation loss for a missing network~\cite{garcia2018modality}.
Finally, Luo \etal~\cite{luo2018graph} learned the direction of distillation between modalities, considering a cosine distillation loss and a representation loss.

Our work differs in three ways:
First, distillation has been applied to either classification or object/activity detection, while our work focuses on transferring knowledge in a control problem by distilling both image representations and RL policies.
Second, distillation has so far been applied from a teacher network to a student network, while we choose to transfer between the auxiliary task (aerial view) and the main task (street view), sharing the local and policy modules in the network.
Third, we employ a novel view dropout to further enhance the transferablity.


 
\subsection{Transfer Learning}
Our work is related to transfer learning~\cite{pan2010survey} in visual domains.
The very basic approach to transfer learning is to pretrain on an existing domain or task and fine-tune on the target ones.
Luo \etal~\cite{luo2018graph} proposed a method to transfer multimodal privileged information across domains for action detection and classification.
Chaplot \etal~\cite{chaplot2016transfer}
studied the effectiveness of pretraining and fine-tuning
for transferring knowledge between various environments for 3D navigation.
Kansky \etal~\cite{kansky2017schema} proposed Schema Networks to transfer experience from one scenario
to other similar scenarios that exhibit repeatable structure and sub-structure.
Bruce \etal~\cite{bruce2017one} leverage an interactive world
model built from a single traversal of the environment, a pretrained visual feature encoder, and stochastic environmental augmentation, to demonstrate successful transfer under real-world environmental variations without fine-tuning.

\begin{figure*}[t]
    \centering
    \includegraphics[width=.85\linewidth]{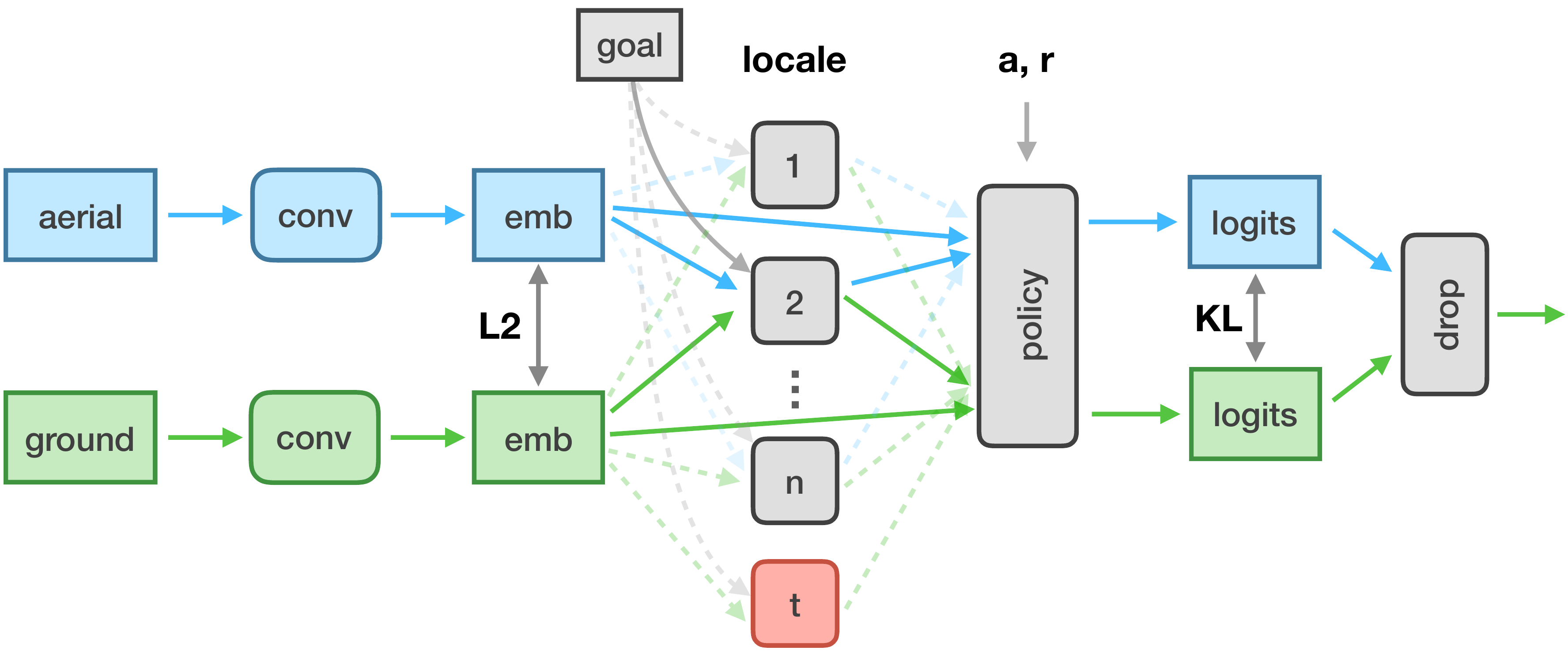}
    \caption{Overview of Cross-view Policy Learning: Ground-view and aerial-view inputs are passed into separate Convolutional Neural Networks for embedding. An $L_2$ embedding loss is used to constrain the similarity between the two latent spaces. The embeddings are passed to a \emph{locale} LSTM (region-specific) and a global policy LSTM (shared across all regions). Both LSTMs are shared across the two views. A KL policy loss is used to constrain the policy logits between the two views. View dropout (gating) selects either of the two views and the final action is sampled according to a multinomial distribution over the logits. This figure shows $n$ regions (gray boxes) for training and one target region (red box) for transfer. Goals are represented by lat/long coordinates. $a,r$ represent the action and reward respectively.}
    \label{fig:overview}
\end{figure*}



\section{Approach: Cross-view Policy Learning}


The full model of our navigation agent is illustrated on Figure \ref{fig:overview}. Both ground-level and aerial view images are fed into the corresponding representation networks, Convolutional Neural Networks (CNN)~\cite{lecun1998gradient} without weight sharing across the two modalities. The image embeddings, output by the CNNs, are then passed into a \emph{locale}-specific LSTM, whose output is then fed into the \emph{policy} LSTM together with the visual embedding. The \emph{policy} LSTM produces logits of a multinomial distribution over actions.
As there are two pathways (for ground-level and aerial views) with two sets of policy logits, an additional gating function decides the final set of logits (either by choosing or merging the two policies) from which to sample the action.


In order to bind the two views and to allow for learning a policy that is interchangeable across views, we proposed to incorporate three ingredients as part of this cross-view policy learning framework: an embedding loss, a policy distillation loss and view dropout, which we detail in the subsequent sections.


\subsection{Reinforcement Learning}
We follow \cite{streetlearn} and employ the policy gradient method for training the navigation agents, learning a policy $\pi$ that maximizes the expected reward $\mathbb{E}[\mathcal{R}]$. 
In this work, we use a variant of the REINFORCE~\cite{williams1992simple} advantage actor-critic algorithm $\mathbb{E}_{a_t\sim\pi_\theta}\left[\sum_t \nabla_{\theta} \log \pi(a_t|s_t, \mathbf{g}; \theta) (\mathcal{R}_t - \mathcal{V}^{\pi}(s_t))\right]$, where $\mathcal{R}_t = \sum_{j=0}^{T-t} \gamma^{j}r_{t+j}$, $r_t$ is the reward at time $t$, $\gamma$ is a discounting factor, and $T$ is the episode length.
In this work, instead of representing the goal $\mathbf{g}$ using distances to pre-determined landmarks, we directly use latitude and longitude coordinates.

We specifically train the agents using IMPALA \cite{impala2018}, a distributed asynchronous actor-critic implementation of RL, with 256 actors for single-region and 512 actors for multi-region experiments, relying on off-policy minibatches re-weighted by importance sampling. Curriculum learning and reward shaping are used in the early stage of agent training to smooth out the learning procedure, similarly to~\cite{streetlearn}.

\begin{figure*}[t]
    \centering
\subfigure[Training]{\includegraphics[width=.25\linewidth]{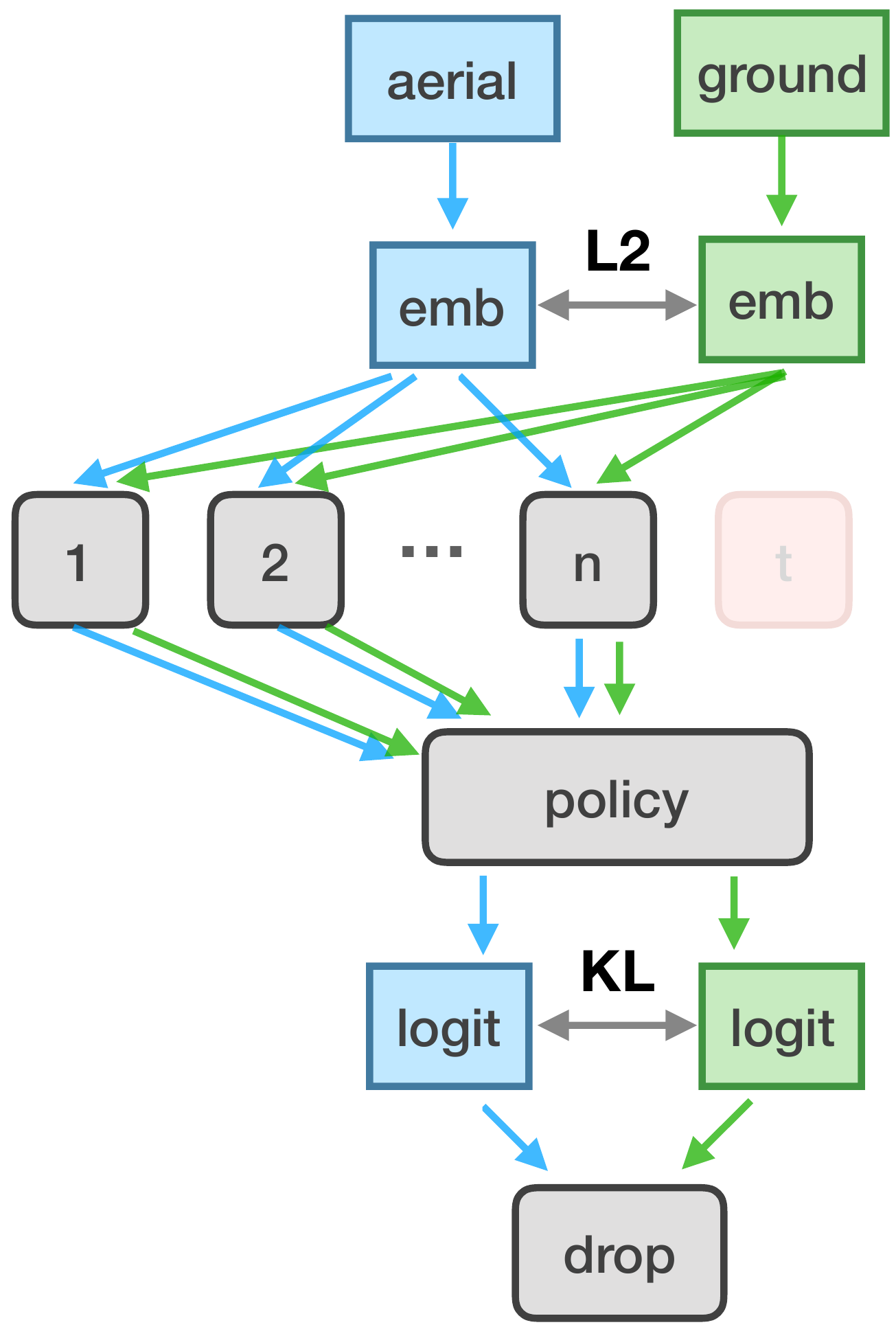}}~~~~~~~~~~~~~
\subfigure[Adaptation]{\includegraphics[width=.25\linewidth]{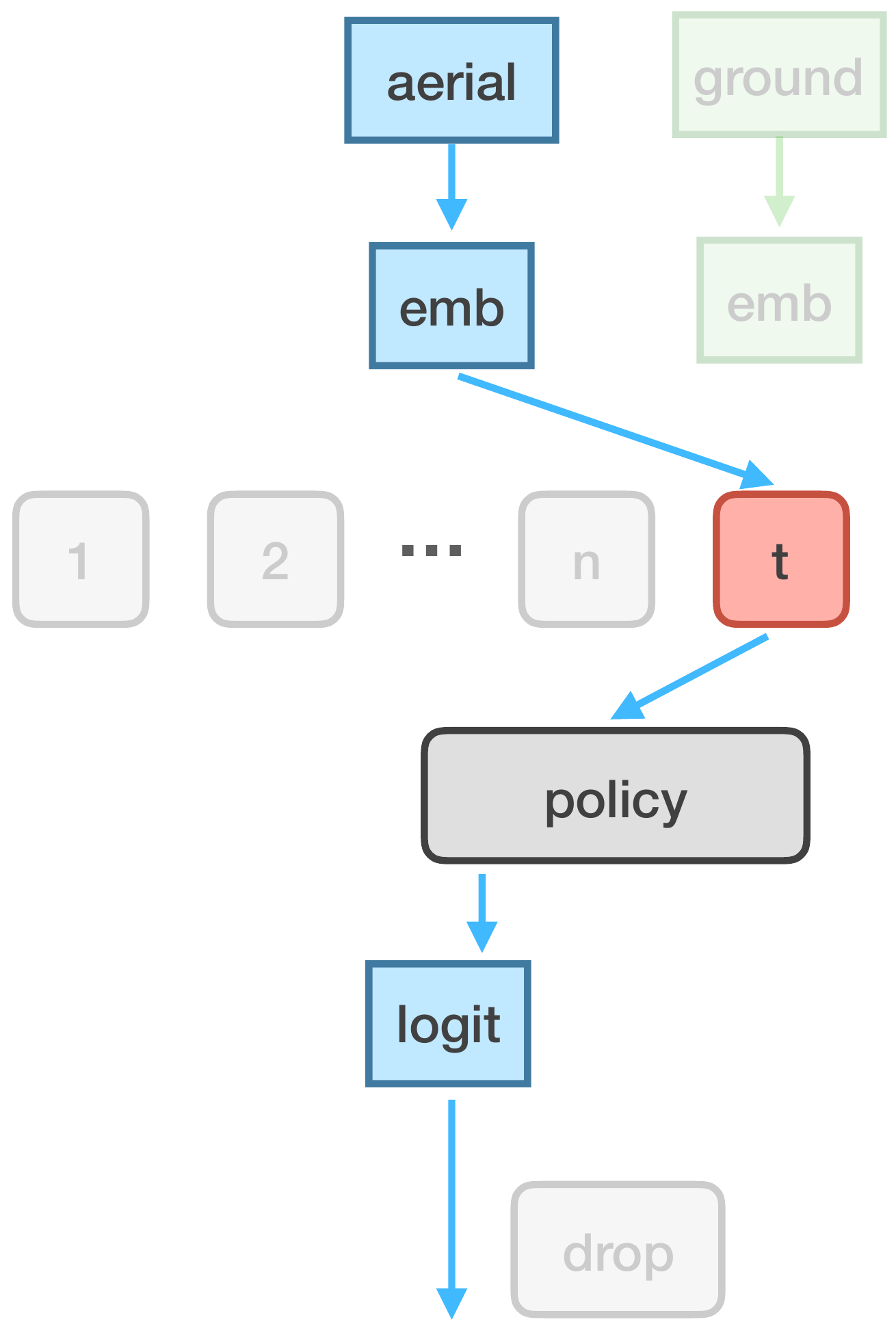}}~~~~~~~~~~~~~
\subfigure[Transfer]{\includegraphics[width=.25\linewidth]{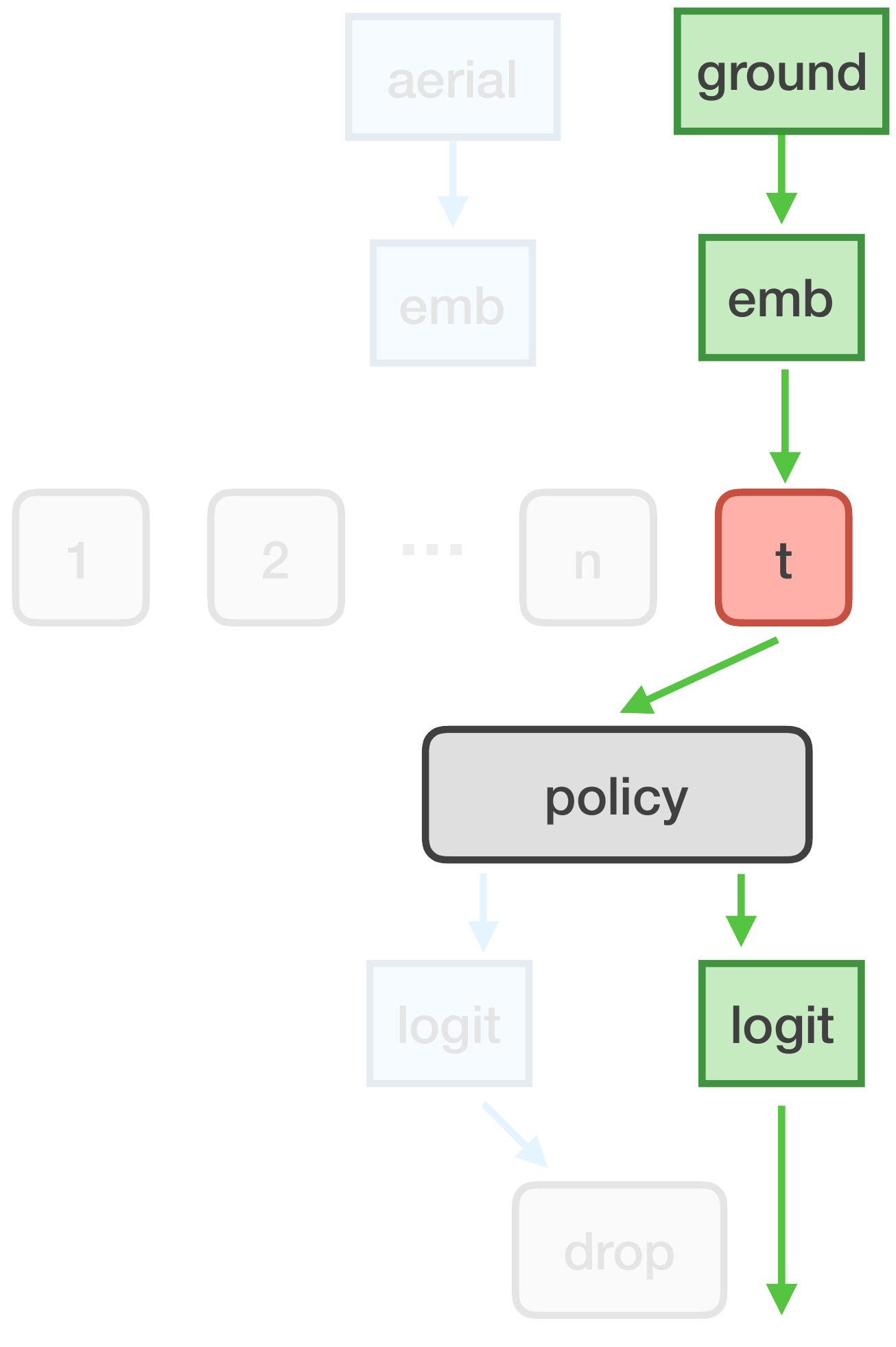}}
\vspace{-1mm}    
    \caption{Transfer learning procedure including 3 stages: training, adaptation and transfer. 
The agent is trained with both ground and aerial view observations in the training city regions. Part of the agent is adapted to the held-out city region, using only aerial view observations. The agent is transferred to the target city region and continuously trained using only ground-view observations.}
    \label{fig:transfer}
\vspace{-4mm}    
\end{figure*}

\subsection{Joint Multi-View Embedding}

There are two reasons why we need to learn a joint representation between the two views in order to exploit the auxiliary aerial view.
First, learning a joint embedding enables us to substitute aerial views for ground-level views at transfer time, once we have adapted the agent to the unseen area using aerial views only.
Secondly, enforcing the embeddings to be similar could potentially make model training faster and more robust. The original representation is only learned through interactions with the environment so ideally such representation should not be dissimilar when one uses signals from different modalities.
Motivated by these, we introduce an embedding loss that enforces learning a joint embedding space between the two views: 
\begin{equation}
     \ell_\text{embed}=\|f_{g}(x_\text{ground})-f_{a}(x_\text{aerial})\|_2~,
\end{equation}
where  $f_g$ and $f_a$ are the CNN modules corresponding to ground-level and aerial view inputs, respectively.

\subsection{Policy Distillation}
\vspace{-1mm}    
Simply minimizing the $L_2$ distance between embeddings may not be sufficient since in practice it is impossible to exactly match one with the other.
The small errors between the two representations could be amplified dramatically as they propagate into the policy networks. So we further propose to match the logits between the policy outputs from the two modalities. In other words, although the embedding between the two modalities may be slightly different, the policy should always try to generate the same actions at the end. Specifically, a Kullback-Leibler divergence loss is added to the total loss, \ie, 
\begin{equation}
    \ell_\text{policy} = -\sum_x p_g(x)\log\left(\frac{p_a(x)}{p_g(x)}\right),
\end{equation}
where, $p_g$ is the softmax output of ground-view policy logits and $p_a$ is the softmax output of aerial-view policy logits. In this way, the learned policy could be less sensitive to differences in representation made by the convolution networks.

\subsection{View Dropout}
\vspace{-2mm}    
While there are two pathways and thus two sets of policy logits, the agent can sample only one action at a time. We propose to fuse the policy outputs of the two modalities through a dropout gating layer, that we call \textit{view dropout} since it chooses over modalities instead of over individual perceptual units. This dropout layer aims at enforcing the cross-modal transferability of the agent.


\subsection{Total Loss Function}
\vspace{-2mm}    
The final objective is
\begin{equation}
    \ell_\text{total} = \ell_\text{RL}+\lambda \ell_\text{embed}+\gamma\ell_\text{policy}
\end{equation}
where $\ell_\text{RL}$ is the reinforcement learning loss. $\lambda$ and $\gamma$ are coefficients indicating the importance of embedding and distillation loss terms respectively. They can be set according to some prior or domain knowledge, or be the subject of hyper-parameter search.

\subsection{Transfer Learning with Cross-View Policy}
\vspace{-2mm}    
We present in this section that a cross-view policy can be used for transfer learning.
Figure \ref{fig:transfer} illustrates the three stages of the transfer learning setting: training, adaptation and transfer. The details of each stage are explained below.

\begin{itemize}[leftmargin=*]
\itemsep 0pt
    \item \textit{Training}: The agent is initially trained on $n$ regions using paired aerial and ground view observations with $L_2$ loss, KL loss and view dropout. All modules (two parallel pathways of CNN, local RNNs and the policy RNN) are trained in this stage.
    
    \item \textit{Adaptation}:
    At the adaptation stage, only the aerial images in the target region are used and only the \emph{locale} LSTM (red box) is trained on the aerial-view environment. Since the ground-level view and the aerial view pathways have been already trained to share similar representations and policy actions, this stage makes the agent ready for substituting the aerial view for the ground-level view during for next phase.
    
    \item \textit{Transfer}: 
    During transfer, the convolution networks and \emph{policy} LSTM of the agent are frozen, with only the target \emph{locale} LSTM being retrained, solely on ground-view observations. The reason why the CNN and \emph{policy} LSTM are frozen is because this modular approach efficiently avoids catastrophic forgetting in already trained city areas (as their corresponding modules are left untouched).
\end{itemize}

\section{Experiments}
In this section, we present our experiments and results, study the effect of curriculum and heading information, perform an ablation study for two components of the loss function, and demonstrate the need for the adaptation stage.

\subsection{Setup}

\noindent\textbf{Goal-Driven Navigation (\emph{Courier} Task).} Following~\cite{streetlearn}, the agent's task consists in reaching, as fast as possible, a goal destination specified as lat/long coordinates, by traversing a Street View graph of panoramic images that cover areas between 2km and 5km a side. Panoramas are spaced by about 10m; the agent is allowed 5 actions: move forward (only if the agent is facing another panorama, otherwise that action is wasted), turn left/right by 22.5 degrees and turn left/right by 67.5 degrees. Upon reaching the goal (within 100m tolerance), the agent receives a reward proportional to the bird flight distance from the starting position to the goal; early rewards are given if the agent is within 200m of the goal. Episodes last for 1000 steps and each time a goal is reached, a new goal location is sampled, encouraging the agent to reach the goals quickly.

\vskip 0.5em\noindent\textbf{Multimodal Egocentric Dataset.}
We build a multiview environment by extending StreetLearn~\cite{streetlearn}. Aerial images are downloaded that cover both New York City and Pittsburgh. At each lat/long coordinate, the environment returns an $84\times84$ aerial image centered at the location, of same size as the ground view image, and rotated according to the agent's heading towards North. Aerial images cover roughly $0.001$ degree spatial differences in latitude and longitude. The training set is composed of four regions: Downtown NYC, Midtown NYC, Allegheny district in Pittsburgh and CMU campus nearby in Pittsburgh, while the testing region is a held-out set and located around the NYU campus and Union Square in NYC, which does not overlap with training areas (see Figure~\ref{fig:my_label} for their approximate locations).

\vskip 0.5em\noindent\textbf{Transfer Learning Setup.}
The real transfer task includes three stages, \ie, training, adaptation and transfer. The agent is trained in one area using both ground-view and aerial-view observations during the training stage with 1 billion steps. In the adaptation stage, the agent only takes in the aerial-view observations and retrains the local LSTM in the target transfer area with 500 million steps. Then the agent navigates in the transfer area with only ground-view observations and is continuously trained. Note that without additional aerial-view observations, an agent cannot be transferred in such a 3-stage setup.

We conduct ablation studies by skipping the adaptation phase (see Section~\ref{sec:ablation}). In that case, the agent is trained on both views in the training regions and learns to navigate in the target region using only ground-view observations. During the transfer stage, it is fine-tuned in the target region.

\vskip 0.5em\noindent\textbf{Architecture.}
Our model is an extension of the model used in \cite{streetlearn} which considered only the ground-view modality. To gain intuition from results effectively, we use the same type of architectures for all networks as in \cite{streetlearn}, and compare our cross-view learning approach with the multi-city navigation agent proposed in \cite{streetlearn} (the latter architecture corresponds to the ground-view pathway in our architecture on Figure~\ref{fig:overview}). 

\vskip 0.5em\noindent\textbf{Parameter Selection.}
As in~\cite{streetlearn}, the batch size is $512$, RMSprop is used with an initial learning rate of $0.001$ and with linear decay; the coefficient of embedding and policy distillation losses were set to $\lambda=1$ and $\gamma=1$. 

\subsection{Cross-View vs. Single-View}
\begin{figure}[!t]
    \centering
    \includegraphics[width=\linewidth]{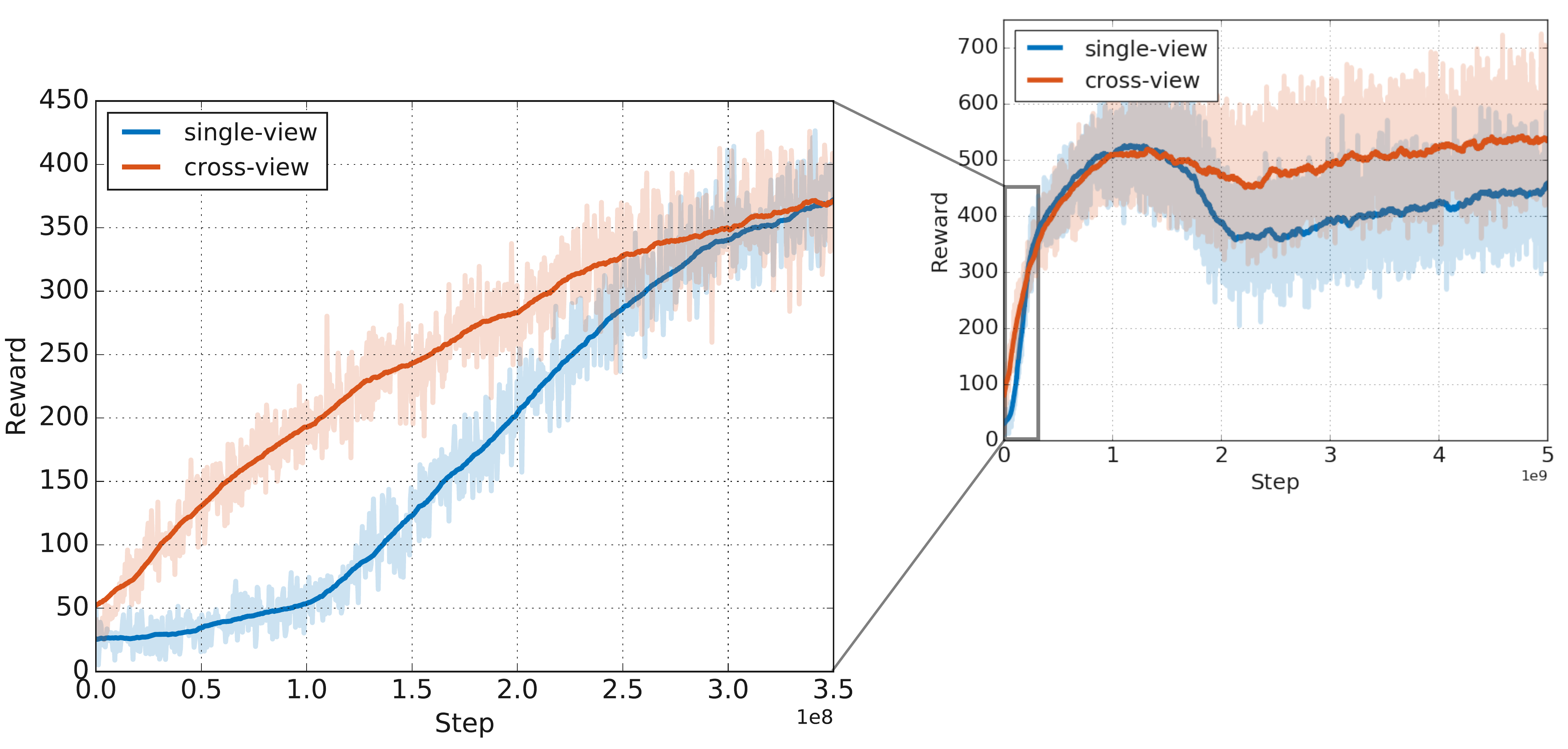}
    \caption{Rewards gained by the agent at the transfer stage in a fixed target city region. The agent is continuously trained during transfer. Higher rewards are better. The proposed cross-view approach significantly outperforms single-view baseline in terms of initial and convergence rewards. The left figure magnifies the rewards within 350M steps, which shows the zero-shot reward and learning speed are both improved significantly.}
    \label{fig:deploy-no-curriculum}
\end{figure}

We start by presenting the rewards in transfer stage gained by the proposed cross-view method and the baseline single-view method in Figure \ref{fig:deploy-no-curriculum}. 
The cross-view agent leveraged the aerial images in the adaptation stage to adapt better to the new environment. However, in transfer stage, both agents only observe the ground-view.
This aligns with real world scenarios well as the top-down aerial-view is not always available in an online manner. 
The \emph{locale} LSTM of the agents are being retrained during the transfer stage; all other components such as CNN and \emph{policy} LSTM are frozen. The target region is fixed and goals are randomly sampled from this region. Heading information is not used since a ``compass'' is not always guaranteed in navigation. 

Figure \ref{fig:deploy-no-curriculum} shows the rewards obtained by cross-view and single-view methods in the transfer phase. We magnify the rewards within 350M steps because we are more interested in early stage performance of transfer learning. The cross-view method achieves around 190 reward at 100M steps and 280 reward at 200M steps, both of which are significantly higher than the single-view method (50 @ 100M and 200 @ 200M). We can see on the figure that the cross-view approach significantly outperforms the single-view method in terms of learning speed at the early stage. 


Besides retraining, we conduct an experiment to evaluate the \textit{zero-shot reward} or \textit{jumpstart reward} \cite{Taylor:2009:TLR:1577069.1755839}, which is obtained by testing the agent in the target region without any additional retraining. The zero-shot reward is averaged over 350M steps. The proposed cross-view method achieves a zero-shot reward of 29, significantly higher than the reward of 5 obtained by the single-view method. We notice that the success rate is non-linearly correlated with the reward. So we also count the corresponding success rate, defined as the number of goals successfully reached within a thousand steps divided by the total number of goals. The cross-view agent achieves $34.5\%$ \textit{zero-shot success rate}, more than 3x the success rate of a single-view agent ($10.5\%$). We attribute this to the adaptation phase using the aerial-view imagery. It is worth noting that the convergence reward of the cross-view method is also significantly higher than that of the single-view method ($580$ vs. $500$) which shows that the cross-view method learns a better representation.

The above results suggest that the proposed transfer learning allows the agent to gain knowledge about the target city region so that the subsequent navigation can start from a good initial status and such knowledge can significantly improve the continual learning of the agents. The results also suggest that the proposed cross-view learning approach is able to significantly improve the generalization of the representation and the transferability of the street-view agent.

\subsection{Curriculum and Heading}
As we mentioned earlier, both the training and adaptation stages utilize a pre-defined curriculum and environment-provided heading information, following~\cite{streetlearn}. The curriculum increases the distance to goals over time; so that the agent always starts from easier tasks (closer to the goals). This time, we incorporate extra heading information during training, by adding an auxiliary supervised task that consists in predicting the heading from observations. Previous transfer experiments did not utilize them because heading may not be available in a real world scenario; in this section, we examine how the curriculum and heading information could affect the performance of the agents. 

\begin{figure}
    \centering
    \includegraphics[width=\linewidth]{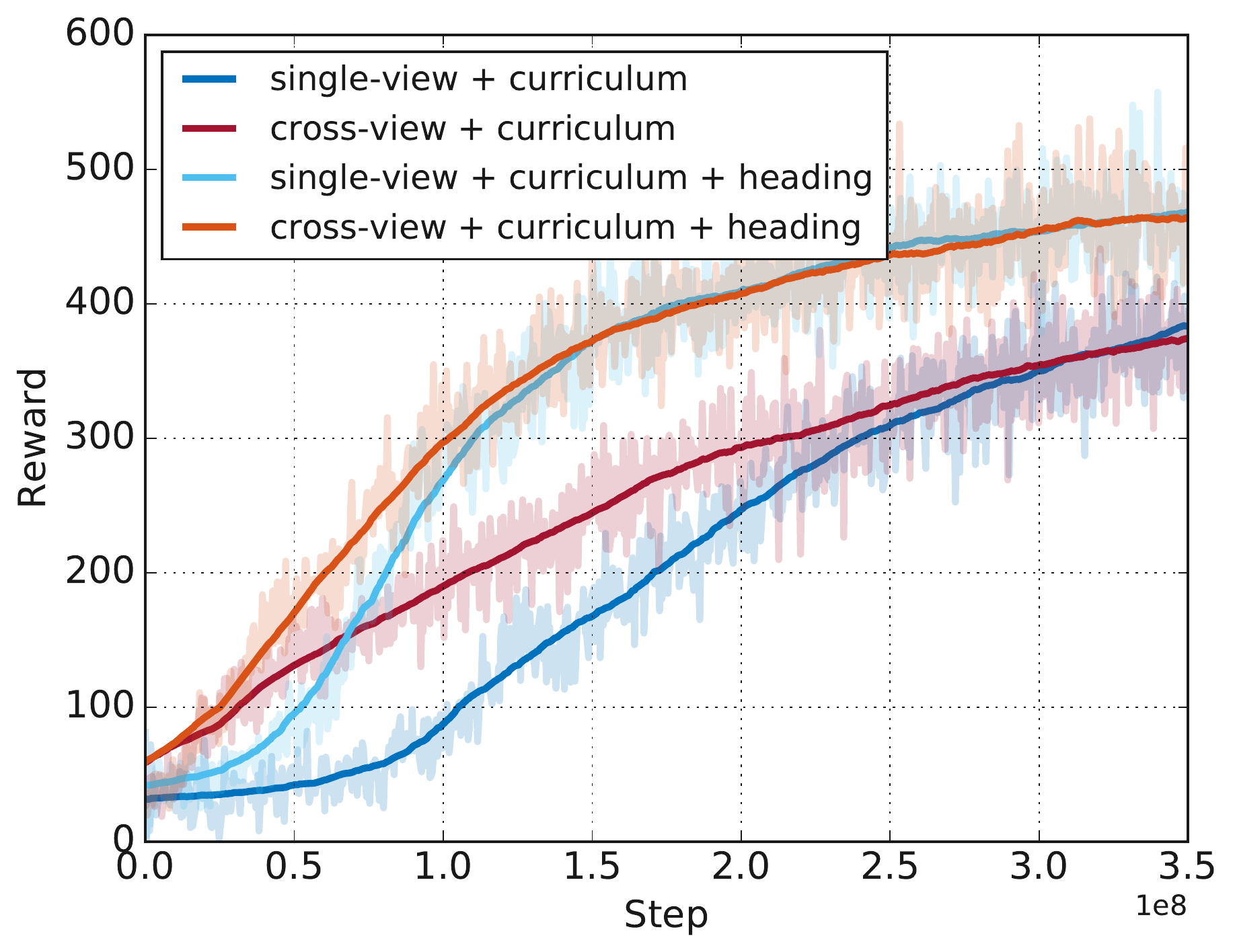}
    \caption{Transfer rewards of agents with curriculum and heading prediction auxiliary task. The performance gap between single-view and cross-view methods are smaller when heading information is used. Heading prediction also leads to higher rewards.}
    \label{fig:curriculum}
\end{figure}

Figure \ref{fig:curriculum} compares transfer phase rewards for four different methods: single/cross views with curriculum, and single/cross views with both curriculum and heading prediction auxiliary tasks. The results suggest that with the heading auxiliary task, the agents can achieve significantly higher performance (approximately 450 reward at step 350M). In addition, the gap between single-view and cross-view is smaller with heading information. 

We also observed that cross-view methods manage to learn irrespective of the curriculum design. In other words, our cross-view architecture compensates for the lack of curriculum by transferring knowledge between cities. Rewards in the cross-view approach grow linearly and reach around 290 at 200M steps, which is comparable with the results shown in Figure \ref{fig:deploy-no-curriculum}. However, the performance of single-view agents degrades significantly without training curriculum. It fails to reach over 50 reward within 100M steps (dark blue curve in Figure \ref{fig:deploy-no-curriculum}), 30 less than the one trained with curriculum (dark blue curve in Figure \ref{fig:curriculum}). Without curriculum learning, the single-view agent learns slowly.

\subsection{Adaptation Using Aerial Views}
\begin{figure}
\centering
    \includegraphics[width=\linewidth]{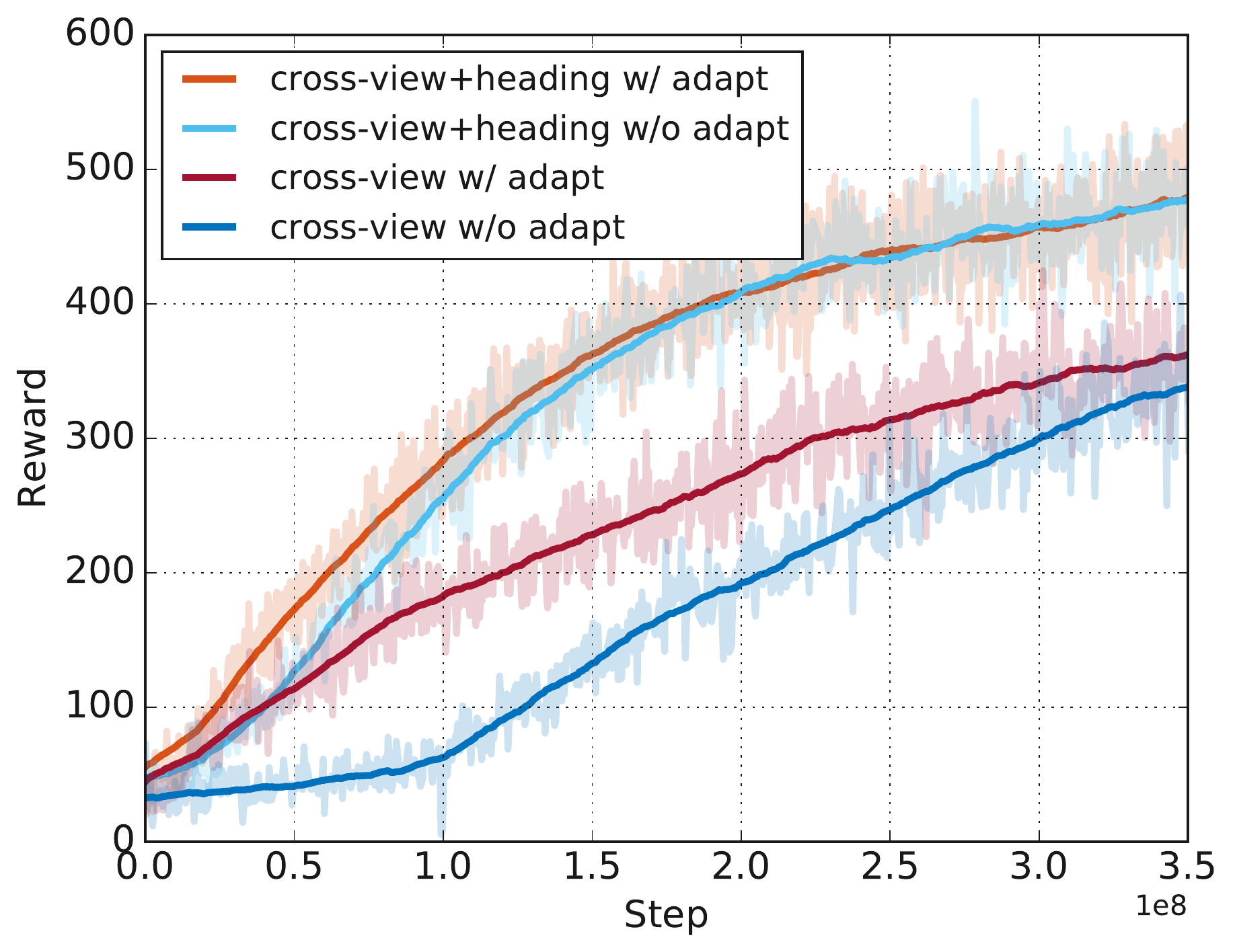}
\caption{Rewards at the transfer phase (with curriculum learning) for cross-view agents going or not through the adaptation stage.}
\label{fig:ablation-transfer}
\end{figure}

An important question is how much improvement is brought by aerial-view based transfer learning. Figure \ref{fig:ablation-transfer} compares transfer phase rewards between 1) cross-view agents that are transferred with aerial-view and 2) agents that skipped the adaptation stage. All transfers are under done using the curriculum. We also compare agents with and without heading prediction.

Figure \ref{fig:ablation-transfer} suggests that the adaptation stage is important and leads to a higher zero-shot reward, faster learning progress in the initial phase and better overall performance of the agent. The effect of adaptation becomes more significant when heading information is dropped during the adaptation stage (which fits better to real world situations). Unsurprisingly, as the agents are fully retrained, their performances become comparable after a sufficiently large number of training steps.

\subsection{Ablation Study}
\label{sec:ablation}
The proposed cross-view policy learning is composed of multiple components: $L_2$ embedding similarity loss, KL policy distillation loss and view dropout. In this section, we evaluate the contribution of each one of those components.

In order to show the strength of view dropout, we implement another approach which uses the same $L_2$ distance loss between embeddings and KL divergence loss between policy logits but always taking the street-view policy logits for action selection (instead of randomly dropping either of the views). In this case, the aerial-view policy logits are not involved in decision making. We name this method ``view distillation'' (in short, \textit{distill}) as an additional baseline since it reflects the setting of model distillation -- one model is optimized for the main objective while the other one is optimized only to match the logits of the former.

\begin{figure}
    \centering
    \includegraphics[width=\linewidth]{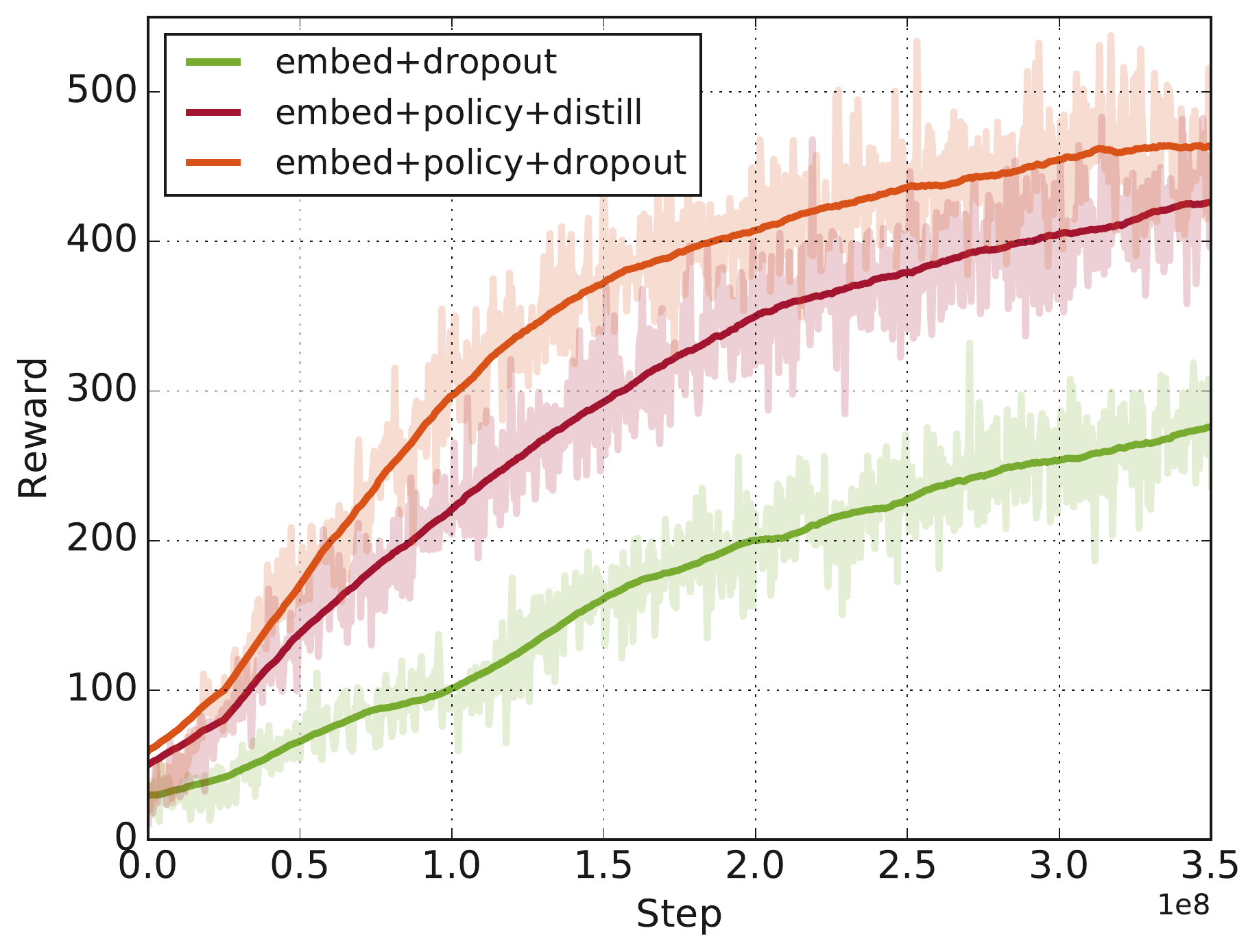}
    \caption{Ablation Study: Transfer the agents under curriculum with heading prediction auxiliary task.}
    \label{fig:ablation-head}
\end{figure}

Figure \ref{fig:ablation-head} shows the rewards for transfer with curriculum and heading auxiliary loss\footnote{The trend for transfer without heading information is very similar.}. Three cross-view methods are compared: (a) full model without KL loss, (b) full model with view dropout replaced with view distillation, and (c) the full model. 

According to the figure, simply using $L_2$ embedding loss without KL policy loss is insufficient to learn a good transferrable representation across views. Its result is significantly worse than the full model. This is probably because the discrepancy between the two views makes it impossible to project them into the same space. There are always differences in their representations and such differences are enlarged after passing through the policy networks. Having an additional KL policy loss would allow the learned policy to be more robust (or less sensitive) to such small differences in feature representations.

One may also notice that the agent (\textit{distill}) that always uses the street-view policy for action selection could achieve decent performance but still is non-trivially worse than the agent that uses view dropout. Such results suggest that the $L_2$ embedding loss and the KL policy loss are able to distill a street-view agent into a good aerial-view agent. However, that distilled policy is not interchangeable across views. Training an agent with view dropout can be seen as replacing the navigation task by a more difficult task where the agent has to learn to quickly switch context at every single step. An agent trained on this harder task generalizes across observation modalities.

\section{Conclusion}
We proposed a generic framework for transfer learning using an auxiliary modality (or view), composed of three stages: (a) training with both modalities, (b) adaptation using an auxiliary modality and (c) transfer using the major modality. We proposed to learn a cross-view policy including learning a joint embedding space, distilling the policy across views and dropping out modalities, in order to learn representations and policies that are inter-changeable across views. We evaluated our approach on a realistic navigation environment, \textit{StreetLearn}, and demonstrated its effectiveness by transferring navigation policies to unseen regions.

One interesting future work would be scaling up the system to cover not only urban areas but also rural areas in different countries. Another extension would consist in providing the agent with the start position in addition to the goal position, so that the problem simplifies to learning to find the optimal path from A to B, without the need for learning to relocalize and to find A. After all, as it happened during the successful journey through unknown seas made by the crew of the \emph{Endurance}, the navigator often knows their starting position, and the interesting question is how to reach the destination.

\paragraph{Acknowledgements.}
We thank Carl Doersch, Raia Hadsell, Karl Moritz Hermann, Mateusz Malinowski, and Andrew Zisserman for their discussions and contributions. 

{\small
\bibliographystyle{ieee_fullname}
\bibliography{main}
}

\end{document}